\documentclass{article}

\usepackage[nonatbib, final]{neurips_2020_tda}

\usepackage[utf8]{inputenc} 
\usepackage[T1]{fontenc}    
\usepackage{amsfonts}       
\usepackage{booktabs}       
\usepackage{hyperref}       
\usepackage{url}            
\usepackage{microtype}      
\usepackage{nicefrac}       
\usepackage{paralist}       
\usepackage{siunitx}        
\usepackage{amsmath}
\usepackage{cmap}
\usepackage{algorithm} 
\usepackage{algpseudocode}
\usepackage{longtable}
\usepackage{caption}
\usepackage{multirow}
\usepackage{booktabs}
\usepackage{ltablex}
\usepackage{lineno}
\usepackage{graphicx}

\title{
  TOTOPO: Classifying univariate and multivariate time series with Topological Data Analysis
}

\author{%
  Polina Pilyugina\\
  Skolkovo Institute of Science and Technology\\
   \And
   Rodrigo Rivera-Castro \\
   Skolkovo Institute of Science and Technology \\
   \AND
   Eugeny Burnaev \\
   Associate Professor,  Skolkovo Institute of Science and Technology\\
}

\begin{document}

\maketitle

\begin{abstract}
  This work is devoted to a comprehensive analysis of topological data analysis for time series classification. 
  Previous works have significant shortcomings, such as lack of large-scale benchmarking or missing state-of-the-art methods. 
  In this work, we propose TOTOPO for extracting topological descriptors from different types of persistence diagrams. 
  The results suggest that TOTOPO significantly outperforms existing baselines in terms of accuracy. 
  TOTOPO is also competitive with the state-of-the-art, being the best on 20\% of univariate and 40\% of multivariate time series datasets. 
  This work validates the hypothesis that TDA-based approaches are robust to small perturbations in data and are useful for cases where periodicity and shape help discriminate between classes.
\end{abstract}

\section{Introduction}\label{intro}
Working with time-series data is a perennial challenge in science.
In medicine, classification of electrocardiogram allows detecting anomalies associated with different illnesses, \cite{Jambukia2015}. 
In ornithology, scientists can classify audio signals of recorded bird sounds for species recognition, \cite{Qian2015}.
These are, however, cases of univariate time series.
The multivariate case remains specially difficult, as it is seen on the established benchmark in the time-series community by \cite{Bagnall2017}.
Compared to univariate time-series, methods for multivariate data remain uncommon.
Similarly, their performance is often underwhelming.

Entries, such as \cite{M2016}, \cite{Nielson2015}, \cite{Anderson2018}, \cite{Rivera-Castro2019-nv}, validate the significance of topological data analysis (TDA) for time series data analysis.
TDA is a set of statistical methods aimed to analyse complex datasets.
It relies on ideas from topology and computational geometry, \cite{Edelsbrunner2002}. 
Especially useful in the case of high-dimensional, complex datasets, as it allows to identify structure and shape. 
This makes TDA robust to small perturbations in data, \cite{Cohen-Steiner2007}. 
Similarly, TDA can extract seasonality and periodicity, \cite{Perea2015}, \cite{Dotko2019}. 
Authors have found success with TDA for a wide array of applications ranging from marketing data, \cite{Rivera-Castro2019}, to cryptocurrencies, \cite{Rivera-Castro2019-lv}. 
This highlights one of its benefits.
It does not require particular assumptions on the data. 
These characteristics make TDA-based models interesting to address the classification of time series.

\paragraph{Innovation} 
This work features several innovations compared to other works, such as \cite{Gidea2018}, \cite{Seversky2016}, \cite{Wu2019}, \cite{Pereira2015}, \cite{Umeda}.
Firstly, we compare our approach not only with baseline models, such as 1-Nearest Neighbour Euclidean or Dynamic Time Wrapping models.
Rather than that, we also compare TOTOPO with other TDA-based approaches.
These techniques are either widely discussed in the literature or are current state-of-the-art models for time series classification.
Secondly, common approaches for application of TDA to time series usually rely on creating point clouds from time series data using sliding windows.
Our approach, on the other hand, additionally includes direct persistence diagrams.
They enable us to enrich the model with additional relevant features.
Finally, we combine models based on different TDA descriptors as an ensemble.
This allows us to enrich TOTOPO even further.
With this, we make it more flexible for different types of signals.

\paragraph{Approach}
In general, our proposed approach consists of four steps. 
First, we create Persistence Diagrams from time series.
We do this both directly and using sliding windows.
Second, we calculate TDA inputs from the PDs.
Third, we train base learners on these TDA inputs as well as the original time series.
Finally, we generate an ensemble of base learners.

\section{Topological ideas behind}
To create the topological inputs, first, we extract persistence diagrams from the original signals in two ways.
We do a direct extraction.
Also, we do a transformation of the time series with a sliding window.
\cite{Cohen-Steiner2007} described how to do a direct extraction of persistence diagrams from signals. 
Such persistence diagrams contain information about extreme points and thus shape of time series. 
Our work is the first in the time-series classification literature to use summaries from these diagrams. 
Further, we transform time series into point clouds using sliding windows embedding and extract PDs from them.
This is akin to the works of \cite{Seversky2016} and \cite{Pereira2015}.

\section{TOTOPO}\label{feats_pds}
In order to include TDA into machine learning pipeline, we create TDA Inputs, which summarize relevant information from PDs. 
In this work, three main sets of TDA Inputs are calculated: Betti series, $L^2$-norms series and TDA Summaries. 

\paragraph{Feature extraction}
For each Persistence Diagram ($PD$), we calculate features separately for the set of 0-dim and 1-dim persistent holes.
We describe the following five features.
First, the number of holes of each dimension $d$.
It is the number of points in the $PD_d$, we define them as $N_d$.
Second, the maximal lifetime of holes of each dimension.
Lifetime is the difference between death and birth, $\max_{d} = \max_{z_i \in PD_d}(d_i - b_i)$.
Third, the number of "relevant" holes of each dimension. This is the number of points with a lifetime higher or equal than $ratio * \max_d$.
$ratio \in (0, 1)$ is the parameter describing the strength of relevancy.
Fourth, the average lifetime of holes of each dimension $mean_d = \frac{1}{N_d} \sum_{i=0}^{N_d}(d_i-b_i)$.
Fifth, the sum of lifetimes of all holes of each dimension $sum_d =  \sum_{i=0}^{N_d}(d_i-b_i)$.
Since direct PDs contain only $0$-dimensional holes, we extract only one set of five features from them.
Overall, TDA Summaries is a set of 15 numerical values.
They summarize information from persistence diagrams. 

\paragraph{Betti series}
Another topological input that we use is Betti series, \cite{Umeda}.
For each dimension, Betti series is a sequence of Betti numbers $B_1, B_2, \ldots, B_k$. 
$k$ is an arbitrary number.
In our case, it is set to 100. 
Each Betti number shows how many persistent holes are "alive" at a particular radius.
Thus, Betti series is a sequence of Betti numbers for each radius.

\paragraph{L2 norms}
\cite{Gidea2018} and \cite{Gidea2018a} discuss the advantages of $L^2$-norms for time series analysis. 
They serve as an inspiration for this work.
Firstly, we extract point clouds $P$ from each time series, using sliding windows of size $d$.
Further, we apply a larger sliding window of size $W$ to vectors from $P$.
It creates a series of small point clouds of size $d \times W$. 
Then for each of these point clouds, we calculate the persistence diagrams and persistence landscapes.
Finally, we extract a series of $L^2$ norms from consequent persistence landscapes.
We can define it as $\|\lambda\|_2 = \bigg( \sum_{k=1}^{\infty} \|\lambda_k\|_2^2 \bigg)^{1/2}$.
Here, $\lambda_k$ are lines in a persistence landscape.

\paragraph{Base learner}
We summarize the architecture of the base learner in \autoref{base_learner}. 
The architecture can process multi-channel input signals or features.
It contains three convolutional blocks, global averaging, and fully connected layer with SoftMax activation in the end. 
As an activation function, we use LeakyReLu. 
Also, we consider dropout for training at the end of two blocks out of three, to avoid overfitting. 
Further, we train the base learners on four sets of inputs: TS — original time series, Betti series, $L^2$-norms, and TDA Summaries.

\paragraph{Ensembling}
The approaches of \cite{Bagnall2017} and \cite{Lines2018} inspire our ensemble.
When we train the base learners, we save their training losses for each dataset.
Then we weight the predictions of each model by its rank among other models. 
We sum all weighted predictions to obtain the final one.
The final prediction is thus the class with a maximal weighted probability.
We outline the ensemble in Algorithm 1.

\begin{algorithm}[ht]
\caption{Ensemble}
\begin{algorithmic}
\State $train\_losses = $ train losses of each model
\State $predictions = $ predictions of each model
\State $model\_names = \{ts,\:Betti,\: TDAFeats, \:L2norms\}$
\State $number\_models = k$
\State $inputs=\{\textbf{ts},\:\textbf{Betti},\:\textbf{TDAFeats}, \:\textbf{L2norms}\}$
	\State Sort ascending $train\_losses$ $\to$ $models\_sorted$
	\State Assign votes: $k, k-1, \ldots, 1$ to best, second best, $\ldots$, worst models. 
    \State $votes = \{models\_sorted_1:k, \: models\_sorted_2:k-1,\: \ldots,  \: models\_sorted_i:1\}$
	        \State $combinated\_prediction = 0$
	        \For{$key$, $vote$ in $votes$}
	            \State $combinated\_prediction \: += predictions\{key\} \times vote$
	        \EndFor
    \State $predicted\_classes = \{argmax(combinated\_prediction_i) \textnormal{ for i in range}(N)\}$
    \State \textbf{return} $predicted\_classes$
\end{algorithmic}
\end{algorithm}

\section{Discussion of results}
\begin{figure}[ht]
\centering
\includegraphics[width=0.9\linewidth]{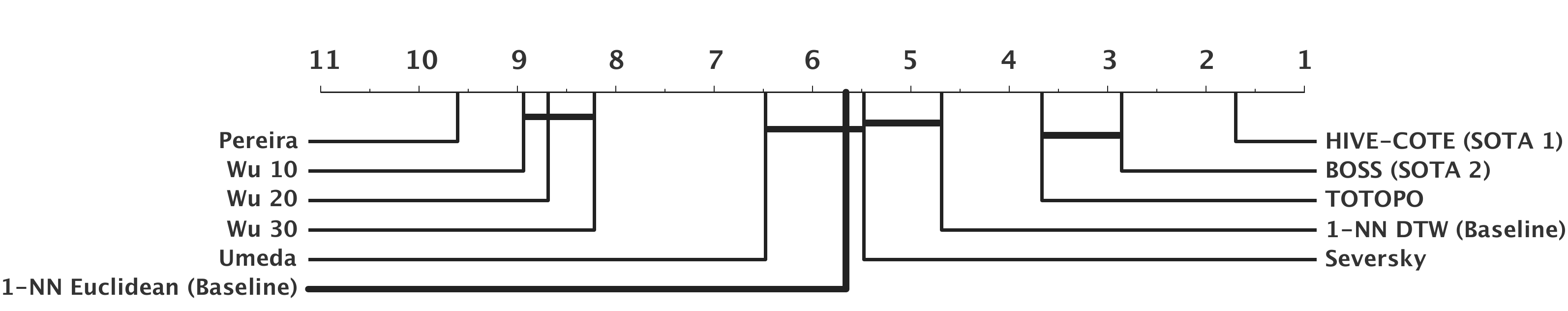}
\caption{CD diagram with black horizontal lines showing statistically similar classifiers.}
\label{cd_diagram}
\end{figure}

We compare TOTOPO against four techniques widely used in the time series community and other TDA-based time series classifiers.
From \cite{Bagnall2017}, we consider 78 datasets for the comparison.
To evaluate the overall performance of algorithms, we use the critical difference (CD) diagram.
We present it in \autoref{cd_diagram}. 
TOTOPO is the third best classifier in the 78 datasets.
It is statistically similar to the state-of-the-art method BOSS.
TOTOPO beats BOSS on 44\% of the datasets.

\paragraph{Multivariate time series}
Multivariate time series methods have less entries in the literature than their univariate counterparts.
Most methods work either for univariate or for multivariate time series.
This is true for the state of the art models for univariate data, such as HIVE-COTE.
TOTOPO has an outstanding characteristic.
For multivariate time series, TOTOPO belongs to the state-of-the-art in terms of performance. 
For this evaluationn, we use the datasets available in \cite{Bagnall2018}.
We can see the results in a comparison against two baselines and a novel SOTA deep-learning architecture by Franceschi, \cite{Franceschi}.
\autoref{tab:multivariate_ranking} presents the average ranks for each model.
Lower values are better.
Our model is the second best performer in the comparison. 

\begin{table}[H]
    \centering
    \caption{Multivariate models, ordered by average rank. Less is better.}
    \begin{tabular}{cccc}
    Method & Average Rank & Method & Average Rank \\ \hline
    Franceschi & 1.79 & 1-NN DTW & 2.58\\
    TOTOPO & 2.5 & 1-NN Euclidean & 3.11\\
    \end{tabular}
    
    \label{tab:multivariate_ranking}
\end{table}

\paragraph{Best results}
We want to highlight the dataset 4-channel ERing.
The data is related to Human Activity Recognition.
On this dataset, our approach significantly outperforms all other methods.
It has an accuracy of 94.4\% as opposed to an average accuracy of 13.3\% in the comparison.
Similarly, TOTOPO excels on the univariate datasets of type DEVICE. 
We show in \autoref{univariate_res} the improvement over Dynamic Time Warping.
Our approach outperforms all methods.
This is the case for both baselines and state-of-the-art, HIVE-COTE and BOSS, on five out of six datasets. 
Time series from this source type correspond to different devices' electricity consumption.
These time series have regions of periodic behaviour.
We assume that TDA makes use of this characteristic.

\paragraph{Limitations}\label{limitations}
TOTOPO underperforms with datasets containing similar classes.
This is the case, for example, for the OliveOil dataset.
The differences between classes among the time series are extremely small.
Our explanation is that TDA is invariant to constant shifts and small perturbations in data.
It requires larger differences in the classes to generate different homologies.

\setlength{\tabcolsep}{2.5pt}
\begin{longtable}{ll|cccccc|c}

\captionsetup{width=15cm}
\caption{Accuracy improvement over DTW across classifiers. Results in \%. More is better.}
\label{univariate_res}\\
\toprule
        & & Seversky & Wu 30 & Pereira & Umeda & HC & BOSS & TOTOPO \\
        Type & Dataset & & & & &  \\
\midrule
\endhead
\midrule
\multicolumn{7}{r}{{Continued on next page}} \\
\midrule
\endfoot
\bottomrule
\endlastfoot
\multirow{6}{*}{\rotatebox[origin=c]{90}{DEVICE}} 
& LargeKitchenAppliances  & -6,93    & -46,14 & 2,94    & -9,33    & 6.93  & -2.94  & \textbf{9.33}\\
& SmallKitchenAppliances  & -21,06  & -50,94  & -8,26    & -21,33  & 21.06  & 8.26  & \textbf{21.33}\\
& ScreenType        & -19,2    & -6,4  & -6,67    & -23,2    & 19.2  & 6.67  & \textbf{23.2}\\
& RefrigerationDevices    & -9,33    & -10,67  & -3,47    & -14,13  & 9.33  & 3.47  & \textbf{14.13}\\
& ElectricDevices      & -16,82  & N/A  & -19,71  & -16,29  & 16.82  & \textbf{19.71}  & 16.29\\
& Computers          & -6    & -20  & -5,59    & -12,8    & 6    & 5.6  & \textbf{12.8}\\
\end{longtable}

\begin{figure}
\centering
\begin{minipage}{.49\textwidth}
  \centering
  \includegraphics[width=\linewidth]{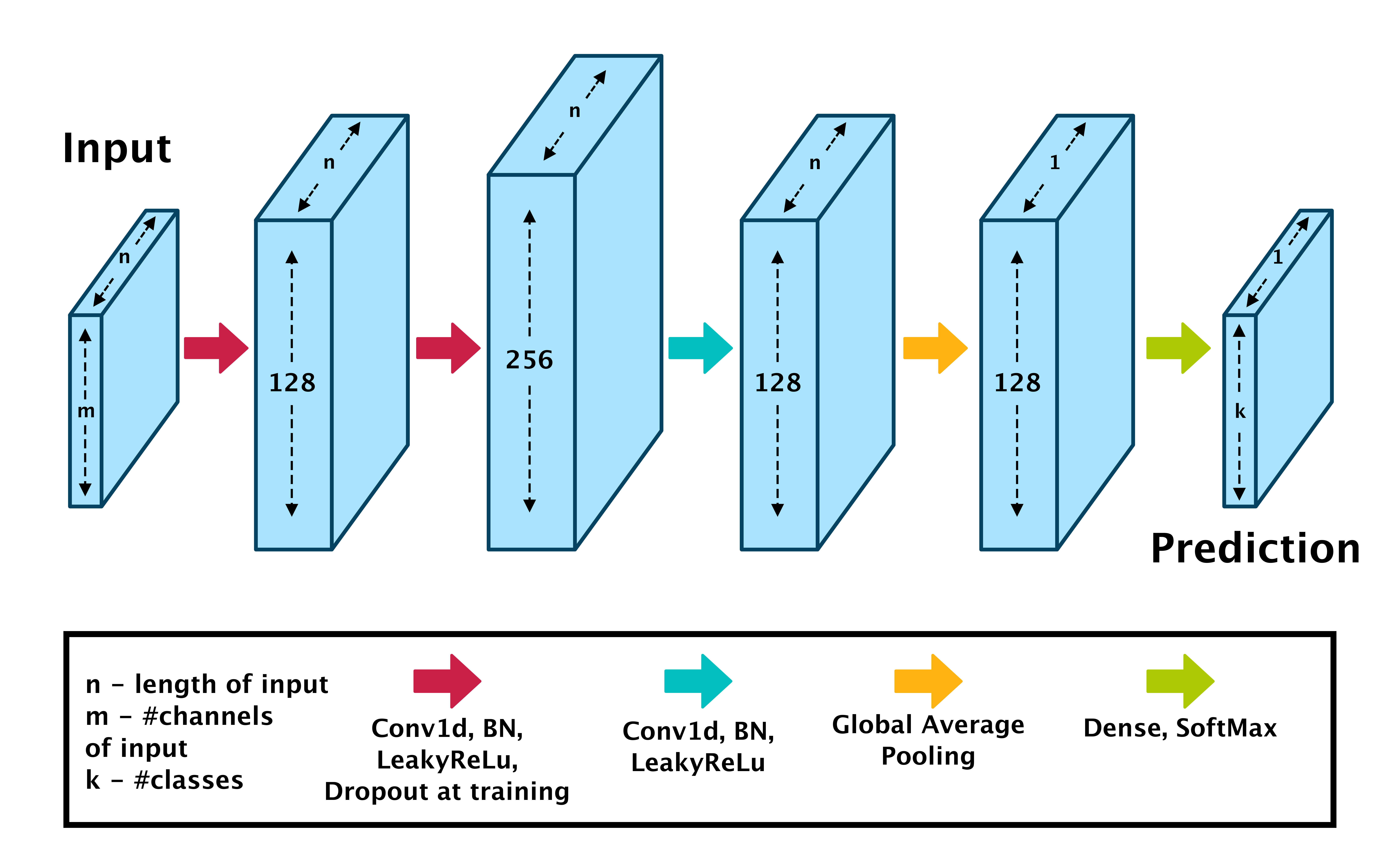}
  \captionof{figure}{Overview of the structure of the base learner classifier}
  \label{base_learner}
\end{minipage}%
\hfill
\begin{minipage}{.49\textwidth}
  \centering
  \includegraphics[width=\linewidth]{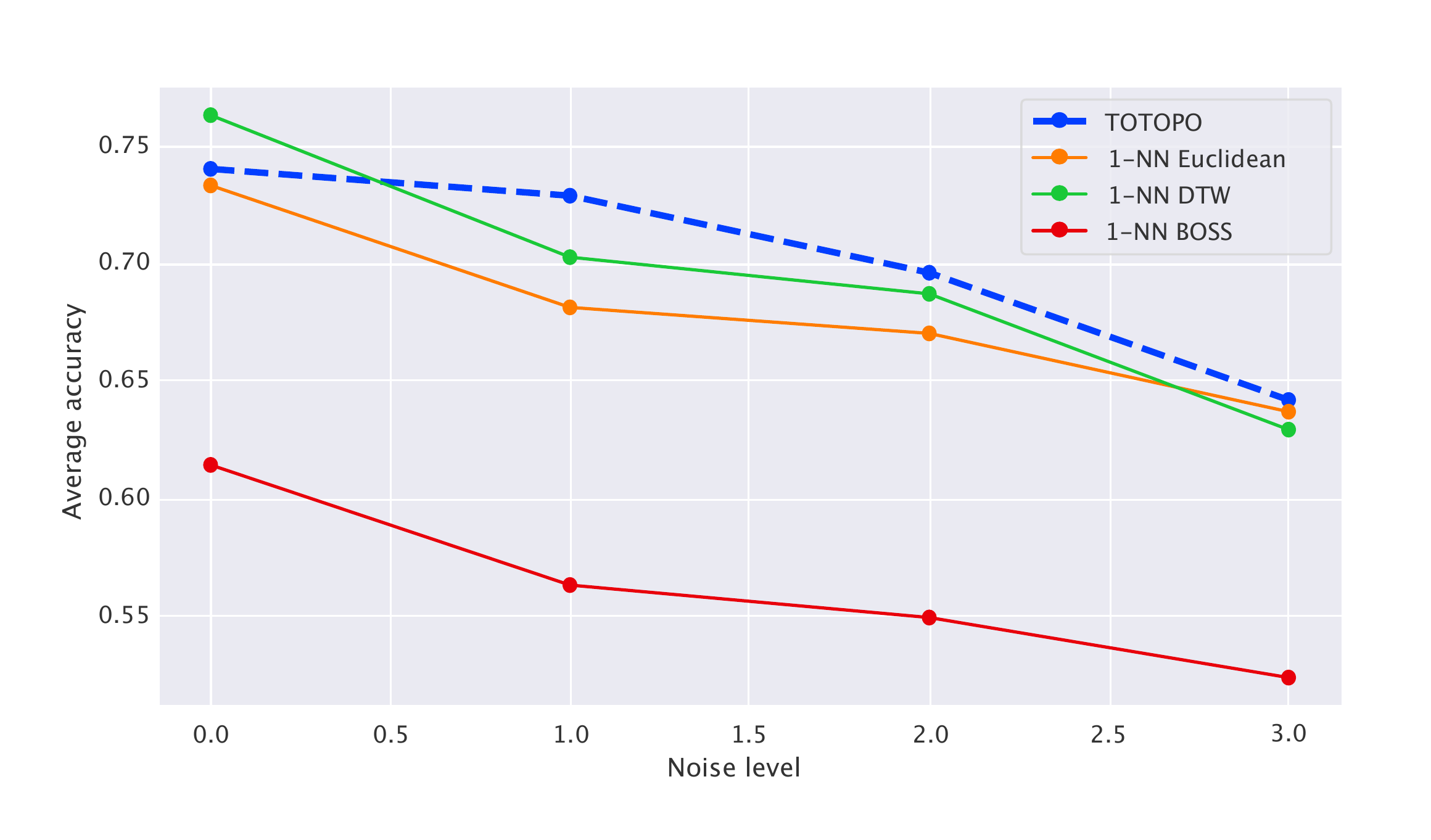}
  \captionof{figure}{Comparison of different algorithms' performances for different noise levels}
  \label{noise}
\end{minipage}
\end{figure}

\paragraph{Noise robustness exploration}\label{noise_robustness}
We want to understand the robustness of TOTOPO.
For this, we apply three noise levels to the signals.
For each noise level we use a targeted Signal to Noise Ration (SNR).
We express it in decibels: 10dB, 15dB and 20dB. 
For each time series $X$, we conver SNR into a linear one.
We follow the suggestions in \cite{Sherman2007}. 
For the comparison, we plot average accuracies for each noise level on \autoref{noise}. 
On the $x$-axis, we show noise levels.
"0" corresponds to the original time series.
"1" is equivalent to an SNR of 20dB, "2" for 15dB and "3" for 10dB.
For small noise levels, TOTOPO outperforms all other models.
Its accuracy deteriorates less.
Overall, this suggests that TDA is indeed robust to noise, especially in case of small changes.

\section{Outlook} 
We aim to improve the base learner classifiers in future works.
This includes the hyperparameter tuning for each of the datasets.
There is also space to optimize the architecture.
Additionally, we seek to evaluate the addition of non-topological descriptors into the model.
With this, we want to overcome the methodological problems of TOTOPO described in \autoref{limitations}.
These can be features to account for constant shifts in data.
We can also use other common descriptors for time series data, i.e., autocorrelation, Fourier coefficients, etc.
Moreover, we aim to create and evaluate new topological descriptors.
They will allow us to improve TOTOPO even further. 

\newpage
\bibliographystyle{abbrv}
\bibliography{main.bib}

\end{document}